%% file: main.tex
\documentclass[letterpaper]{article} 
\usepackage{aaai2026}
\usepackage{times}  
\usepackage{helvet}  
\usepackage{courier}  
\usepackage[hyphens]{url}  
\usepackage{graphicx} 
\usepackage{natbib}  
\usepackage{xspace}
\usepackage[linesnumbered,ruled,vlined]{algorithm2e}
\usepackage{caption} 
\usepackage{amsmath} 
\usepackage{subcaption}
\usepackage{amsfonts}
\usepackage{multirow}
\usepackage{colortbl}
\usepackage{graphicx}
\usepackage{float}
\usepackage{colortbl}
\usepackage[table]{xcolor}
\usepackage{relsize}
\usepackage{newfloat}
\usepackage{listings}
\DeclareCaptionStyle{ruled}
{labelfont=normalfont,labelsep=colon,strut=off} 
\floatstyle{ruled}
\newfloat{listing}{tb}{lst}{}
\floatname{listing}{Listing}
\nocopyright 

\input{preamble}
\title{\method: \underline{S}patially-Aware \underline{I}mage \underline{F}ocus for Visual Reasoning}

\author {
    Zhangquan Chen\textsuperscript{\rm 1}\thanks{The work was conducted during the internship of Zhangquan Chen at ByteDance.},
    Ruihui Zhao\textsuperscript{\rm 3},
    Chuwei Luo\textsuperscript{\rm 3},
    Mingze Sun\textsuperscript{\rm 1},
    Xinlei Yu\textsuperscript{\rm 4},\\
    Yangyang Kang\textsuperscript{\rm 2,3}\footnotemark[2],
    Ruqi Huang\textsuperscript{\rm 1}\thanks{Corresponding authors: ruqihuang@sz.tsinghua.edu.cn, yangyangkang@bytedance.com}
}

\affiliations {
    \textsuperscript{\rm 1}Tsinghua Shenzhen International Graduate School, Tsinghua University, China\\
    \textsuperscript{\rm 2}Zhejiang University, China   \textsuperscript{\rm 3}ByteDance, China\\
    \textsuperscript{\rm 4}National University of Singapore, Singapore
}

\usepackage{bibentry}

\begin{document}
\maketitle
\input{Sec/abstract}
\input{Sec/intro}

\input{Sec/related}

\input{Sec/method}
\input{Sec/exp}
\input{Sec/conclusion}
\small
\bibliography{aaai2026}

\clearpage
\newpage
\appendix
\input{Sec/supmat}

\end{document}

%% file: preamble.tex
\newcommand{\method}{\textit{SIFThinker}\xspace}

%% file: Sec/abstract.tex
\begin{abstract}


Current multimodal large language models (MLLMs) still face significant challenges in complex visual tasks (e.g., spatial understanding, fine-grained perception). Prior methods have tried to incorporate visual reasoning, however, they fail to leverage attention correction with spatial cues to iteratively refine their focus on prompt-relevant regions.
In this paper, we introduce \method, a spatially-aware “think-with-images” framework that mimics human visual perception. Specifically, \method enables attention correcting and image region focusing by interleaving depth-enhanced bounding boxes and natural language. Our contributions are twofold: 
First, we introduce a reverse-expansion-forward-inference strategy that facilitates the generation of interleaved image-text chains of thought for process-level supervision, which in turn leads to the construction of the \textbf{SIF-50K} dataset.
Besides, we propose \textbf{GRPO-SIF}, a reinforced training paradigm that integrates depth-informed visual grounding into a unified reasoning pipeline, teaching the model to dynamically correct and focus on prompt-relevant regions. Extensive experiments demonstrate that \method outperforms state-of-the-art methods in spatial understanding and fine-grained visual perception, while maintaining strong general capabilities, highlighting the effectiveness of our method. Code: \url{https://github.com/zhangquanchen/SIFThinker}. 
\end{abstract}

%% file: Sec/intro.tex
\section{Introduction}
\input{Figs/pipeline}
\label{sec:intro}
Visual understanding is a fundamental task in computer vision, enabling machines to perceive, interpret, and interact with their surroundings~\cite{zeng2025FSDrive, wolfe2017five,intro2:guo2016deep,intro2:palmeri2004visual, zeng2025janusvln, Wang2025OneImage, li2023hong, lee2025skyfall}. Traditional methods typically process the entire image in a uniform manner~\cite{intro:chen2024sharegpt4v,intro:chen2025sharegpt4video,intro:jiao2025lumen,intro:shao2024lmdrive,intro:zhang2024eventhallusion}, without considering the dynamic attention shifts and the underlying spatial awareness. However, an RGB image is inherently the 2D projection of 3D world, and human perception is dynamic attention shifts rooted in 3D awareness.
For example, when prompted to identify the color of shirt worn by the person behind the tree, humans do not perceive the scene in a single step. They first locate a coarse region of interest within their field of view. Then, they progressively focus on the tree while considering its 3D spatial relationship to the surroundings in mind. Finally, they correct their attention to the area behind the tree to determine the color of the shirt worn by the person. 

The above observations indeed reveal two critical points in designing a human-like visual understanding framework -- 1) \textbf{Dynamic visual perception}, which enables attention correction and facilitates focusing on prompt-relevant regions throughout the reasoning process; and 2) \textbf{3D spatial awareness}, which grounds visual perception within a spatial context. 
Moreover, \emph{these two points can be naturally integrated into a unified framework}.


For the former, dynamic visual perception was initially pursued through reasoning frameworks. Early efforts, such as V*~\cite{rela:vstar}, VisCoT~\cite{shao2024visual}, and VisRL~\cite{chen2025visrl}, incorporated visual inputs into the reasoning paradigm in a stepwise manner: first predicting bounding boxes, then performing further inference based on the cropped image. This kind of fragmented method often severs the continuity of the reasoning chain, leading to weaker interpretability and incoherent anchoring of visual regions. 
More recent think-with-images methods, such as Cogcom~\cite{qi2024cogcom} and ChatGPT-o3~\cite{openai2025o3o4mini}, utilize specialized-model-based or tool-based adaptive zooming to simulate dynamic attention. However, these methods are not intrinsic and heavily rely on external capabilities. 
In contrast, \emph{\method intrinsically supports coherent image-text interleaved reasoning by simulating human-like visual attention correction and focusing}.

For the latter, 
the limitation primarily arises from the insufficient modeling of spatial information in MLLMs, i.e., most of them are pre-trained on RGB images paired with textual data without explicit spatial cues. 
Humans, on the other hand, are often capable of inferring 3D relationships from a single 2D image, enabling a deeper understanding of visual scenes.
Prior works such as SpatialBot~\cite{cai2024spatialbot} and SSR~\cite{liu2025ssr} attempt to equip models with 3D perception via data-driven or tool-based approaches.
However, In real-world interactive scenarios, attention is seldom global. In contrast, \emph{we typically perceive the spatial information (e.g., depth) of the prompt-relevant regions rather than the entire image}.

Moreover, previous works often fail to \emph{integrate fine-grained visual perception with spatial awareness}, and typically separate visual grounding from answer generation (e.g., VLM-R1~\cite{shen2025vlmr1}). In contrast, we argue that 3D perception serves as a fundamental basis for deeper visual understanding. Besides, grounding-conditioned reasoning facilitates more accurate answer derivation, while result-level supervision can guide the correction of bounding boxes. 
Thus, there is a need for a spatially-aware, visually grounded reasoning process that offers a more coherent, unified, and effective framework for visual understanding.

To this end, we propose an adaptive image focus mechanism with 3D awareness, enabling the model to selectively and dynamically perceive the focused region with depth throughout the reasoning process. That is, we \emph{integrate visual perception and spatial awareness into a unified grounded reasoning pipeline}. 

Our method consists of three components: data generation, warm-up, and reinforcement learning (RL). In the data generation stage, we propose a novel image–text interleaved chain-of-thought (CoT) generation scheme that simulates spatially-aware visual focus. The constructed data support the warm-up stage via supervised fine-tuning (SFT), guiding the model to adopt structured interleaved reasoning. In the reinforcement learning stage, we go beyond outcome-only rewards and integrate format, depth estimation, region grounding, and answer prediction for each rollout.

Our main contributions can be summarized as follows.

\begin{itemize}
\item We present \method, the first framework to incorporate an adaptive focus mechanism with 3D awareness, enabling spatially grounded visual reasoning.
\item We introduce a novel reverse-expansion–forward-inference strategy for constructing image-text interleaved CoTs, and release the \textbf{SIF-50K} dataset for process-level supervision.
\item We propose \textbf{GPRO-SIF}, which applies reinforcement learning for spatially-aware image focus training: 1) for visual grounding, we propose $HIoU$, a hierarchical design that better accommodates multiple objects, along with an effective reward formulation for bounding box correction; 2) for spatial awareness, we embed depth estimation into the model’s autoregressive structure; 3) we further design a progressive reward to encourage performance improvement across training epochs.
\item Extensive experiments demonstrate that \method outperforms prior state-of-the-art methods in both fine-grained visual perception and spatial intelligence, while maintaining stable general capabilities—highlighting the robustness and broad applicability of our method.
\end{itemize}

%% file: Figs/pipeline.tex
\begin{figure*}[]
    \centering
    \vspace{-2em}
    \includegraphics[width=\textwidth]{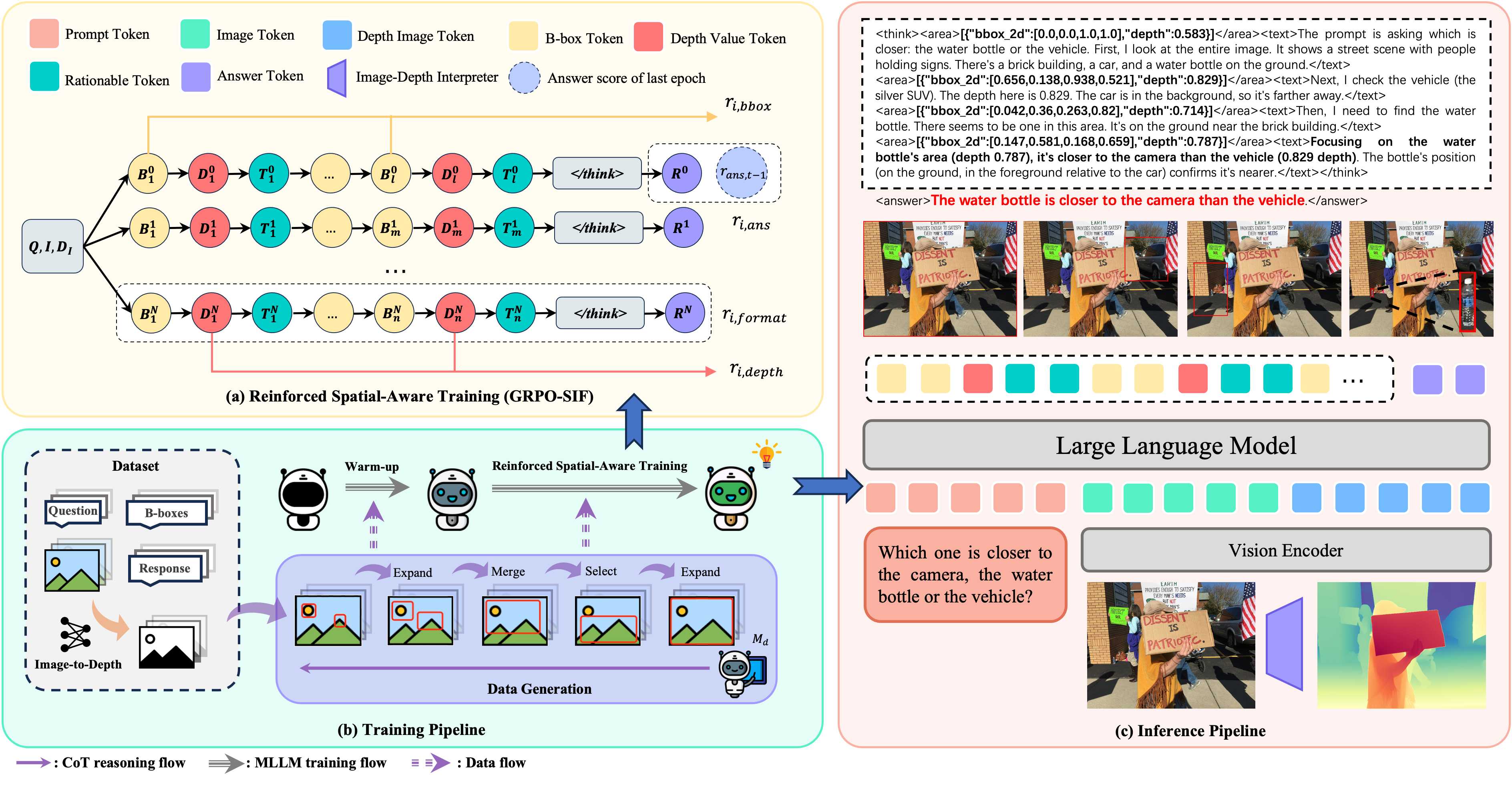}
    \vspace{-2em}
    \caption{The schematic illustration of \method. (a) We propose a spatially-aware image focus paradigm, in which four novel reward functions are introduced under the RL framework to guide the optimization process. (b) The training pipeline of \method is illustrated, which builds upon our proposed data generation pipeline. It begins with a warm-up stage, followed by GRPO-SIF as described in (a). (c) illustrates the inference pipeline of our method given a question-image input.}\label{fig:pieline}
    \vspace{-1.6em}
\end{figure*}

%% file: Sec/related.tex
\section{Related Work}
\label{sec:related}
\subsection{Visual Chain-of-Thought Reasoning}
Recent studies have shown that step-by-step reasoning through in-context learning can significantly enhance the performance of large language models (LLMs). Accordingly, several approaches have emerged that aim to strengthen the visual reasoning capabilities of multi-modal large language models by introducing chain-of-thought strategies. These approaches can be categorized into three types: (T1) Pure-Text-Thought Methods: ~\cite{rela_new:r1v, rela_new:llamavo1, rela_new:alignanything, rela_new:openrlhf, shen2025vlmr1,univg-r1} elicit purely textual CoT reasoning within MLLMs for visual reasoning tasks inspired by \cite{rela:deepseekr1}. They employ reinforcement learning to guide the generation process towards final answers without explicitly incorporating intermediate visual signals. (T2) Intermediate-Thought Methods: ~\cite{rela:liu2025spatialcot, shao2024visual, chen2025visrl, rela:wang2024segllm} first generate fine-grained visual cues (e.g, bounding boxes, spatial coordinates, or segmentation masks), CoT reasoning is then conducted based on these fine-grained visual cues. (T3) Multi-Modal-Thought Methods: Some recent methods aim to integrate visual-textual reasoning more tightly into the model’s thought process. Proprietary systems such as ChatGPT-o3~\cite{openai2025o3o4mini} demonstrate the ability to "think-with-images" by dynamically invoking external image tools. Similarly, ~\cite{rela:mvot} enables visual thinking by generating visual traces of reasoning. ~\cite{su2025openthinkimg, wu2025vtool, zheng2025deepeyes} optimize tool-usage ability via reinforcement learning. Moreover, ~\cite{zhang2025chainoffocus} iteratively crops the image based on generated bounding boxes to extract new visual cues for reasoning. ~\cite{fan2025grit} takes a more direct approach by generating reasoning chains that interleave natural language and explicit bounding boxes.

However, existing methods still exhibit several limitations: (T1) they rely heavily on textual reasoning, neglecting the dynamic visual attention shifts during the reasoning process; (T2) they produce less interpretable and often incoherent reasoning chains; and (T3) Some depend on external tools, specialized detection models or unstable image generation, while others overlook intermediate visual signals, relying solely on outcome-based supervision. Therefore, there is a need for an adaptive and coherent intrinsic method that enables visual-grounded reasoning—allowing MLLMs not only think "about" images, but also \emph{dynamically focus, and correct their visual attention across image regions in a human-like manner.}

\subsection{Spatial Intelligence}
Large language models have undergone swift evolution, exhibiting robust capabilities across a broad spectrum of tasks~\cite{wen2025optimizationbidirectionalgatedloop, zeng2025bridgingeditinggapllms, Leo2024ICAICA, ni2025recondreamer, wang2025embodiedreamer, xiang2025promptsculptor, xu2025finmultitime, zhang2025sensitivitylora, chen2024three}. However, existing MLLMs~\cite{wu2024next, driess2023palm, li2023blip, chen2022pali} are primarily trained on RGB images paired with textual data, which inherently lacks 3D spatial information. As a result, they demonstrate limited performance on tasks requiring spatial reasoning. To address this limitation, recent efforts such as SpatialRGPT
~\cite{cheng2024spatialrgpt} and SpatialVLM~\cite{chen2024spatialvlm} have enhanced the spatial reasoning capabilities of MLLMs by constructing specialized spatially-oriented question-answering datasets and fine-tuning models accordingly. To further emphasize integrated reasoning capabilities, SSR~\cite{liu2025ssr} incorporates depth images as additional inputs, while SpatialBot~\cite{cai2024spatialbot} leverages depth estimation tools to acquire spatial priors in key perceptual regions. However, these spatial perception approaches focus solely on reasoning, without achieving a deep integration with visual grounding—\emph{two processes that are fundamentally interdependent in human visual perception.}

%% file: Sec/method.tex
\section{Methodology}
As shown in Fig.~\ref{fig:pieline}, \method incorporates depth-enhanced focused regions of the image into the thinking process, enabling spatially grounded visual focus. \method can iteratively analyze and refine the regions of interest, ultimately delivering a more accurate final response. In the following sections, we provide a detailed description of the data generation pipeline, the spatially-aware image focus training paradigm, and the \textbf{GRPO-SIF}.
\input{Sec/algo}

\label{sec:method}
\subsection{Data Generation}\label{sec: data-generation}
To simulate the way humans observe spatial scenes, we model a focus mechanism that incorporates depth for data generation. That is, we construct \textbf{SIF-50K}, a dataset consisting of two parts: (1) a tailored fine-grained reasoning subset derived from spatial scenes in Flickr30k~\cite{flickr30k}, Visual7W~\cite{visual7w}, GQA~\cite{gqa}, Open Images~\cite{openimages}, VSR~\cite{vsr}, and Birds-200-2021~\cite{cub}, based on VisCoT~\cite{shao2024visual}; and (2) a multi-instances subset resampled from TallyQA~\cite{acharya2019tallyqa}. All source datasets include ground-truth bounding boxes (b-boxes) annotations. 

As illustrated in Alg.~\ref{alg:reverse-expansion}, given each question-image-b-boxes-answer pairs $(Q, I, B_{gt}, R)$, we apply a reverse expansion procedure, followed by forward reasoning over the expanded regions based on DepthAnythingV2~\cite{yang2024depth} and Doubao-1.5-vision-pro~\cite{guo2025seed1}. This process yields the final SIF-50K dataset, denoted as $\mathcal{P} = \{(Q, I, D_{I}, B_{gt}, R, R_{cot})\}$.

\subsection{Spatially-aware image focus training paradigm}
\paragraph{Method overview.} We propose a two-stage pipeline to incorporate spatially-aware grounded reasoning. The first stage is a warm-start supervised fine-tuning phase, which biases the model to generate structured reasoning chains with explicitly focused regions, resulting in $\mathcal{M}_{SFT}$. This is followed by a reinforcement learning phase that further refines and optimizes these grounded behaviors, yielding the final model $\mathcal{M}_{RL}$. For SFT, we utilize the full set from SIF-50K to get $\mathcal{P}_{SFT} = {(Q, I, D_{I}, R_{cot})}$. For RL (detailed in the following section), to promote progressive learning with minimal supervision, we sample 200 instances from SIF-50K to form a smaller set $\mathcal{P}_{RL} = {(Q, I, D_{I}, B_{gt}, R)}$.

\subsection{Reinforcement Learning with GRPO-SIF}
\paragraph{RL Formulation.} Built upon the Group-Relative Policy Optimisation (GRPO)~\cite{shao2024deepseekmath}, the model $\mathcal{M}_{SFT}$ is framed as a policy $\pi_\theta$ that generates an output sequence conditioned on the input $(Q, I, D_{I})$. During training, for each question-image-depthimage pair $(Q, I, D_{I})$, GRPO-SIF sample a set of $N$ candidate completions $\{o_1, \ldots, o_N\}$ from the current policy $\pi_{\theta_\text{old}}$, and then updates the current policy $\pi_\theta$ by maximizing the following objective:

\begin{align}
\mathcal{J}(\theta) = &
\frac{1}{N} \sum_{i=1}^N \frac{1}{|o_i|} \sum_{t=1}^{|o_i|} \Big\{ \nonumber \min \Big[
\text{clip}(r_{i,t}, 1 - \epsilon, 1 + \epsilon)\hat{A}_{i,t},\; 
\\&r_{i,t} \hat{A}_{i,t}\Big] \nonumber - \beta\, \mathbb{D}_\text{KL}[\pi_\theta \,||\, \pi_\text{ref}]\Big\},
\end{align}

where $r_{i,t} = \frac{\pi_\theta(o_{i,t} \mid q, o_{i,<t})}{\pi_{\theta_\text{old}}(o_{i,t} \mid q, o_{i,<t})}$ denotes the ratio between the new and old policies at step $t$, $\epsilon$ and $\beta$ are hyperparameters. $\mathbb{D}_\text{KL}[\pi_\theta \,||\, \pi_\text{ref}]$ estimate the KL divergence with the unbiased estimator~\cite{schulman2020approximating} between current policy model and the reference model.
For each completion $o_i$, a task-specific reward $r_{i,t} = R(Q, I, D_{I}, B_{gt}, R, o_i)$ is computed based on a combination of reward components (detailed below) at step $t$. These rewards are then used to compute a group-normalized advantage:
\begin{equation}
\hat{A}_{i,t} = \frac{r_{i,t} - \text{mean}\{r_{1,t}, \ldots, r_{N,t}\}}{\text{std}\{r_{i,t}, \ldots, r_{N,t}\} + \delta}.
\end{equation}

$\delta$ is a small constant (e.g., $10^{-8}$) added for numerical stability. The task reward $r_{i,t}$ is a composite signal comprising four components: a spatially-aware grounded reasoning-format reward ($r_\text{format}$), a progressive-answer-accuracy reward ($r_{\text{ans},t}$), a correction-enhanced-grounding reward ($r_\text{bbox}$), and depth-consistency reward ($r_\text{depth}$). These components are designed to jointly encourage spatially-aware grounded reasoning, thus attaching the precise answer.

\paragraph{Hierarchical Intersection over Union ($HIoU$).}
To comprehensively reward the grounding quality between predicted b-boxes $B_p = \{b_1^p, b_2^p, \dots, b_n^p\}$ and ground-truth b-boxes $B_{gt} = \{b_1^{gt}, b_2^{gt}, \dots, b_m^{gt}\}$, we propose the Hierarchical IoU ($HIoU$) as shown in Fig.~\ref{fig:hiou}. This design mitigates reward hacking issues such as inflated AP reward scores by incorporating both global and instance-level complementary components. (1) We first compute the global IoU ($GIoU$), which quantifies the overall spatial consistency between the predicted b-boxes and ground-truth b-boxes as a whole:
\begin{equation}
GIoU = 
\frac{\left| \bigcup_{b_i^p \in B_p} b_i^p \cap \bigcup_{b_j^{gt} \in B_{gt}} b_j^{gt} \right|}{\left| \bigcup_{b_i^p \in B_p} b_i^p \cup \bigcup_{b_j^{gt} \in B_{gt}} b_j^{gt} \right|}.
\end{equation}
(2) Next, we compute a pairwise IoU via one-to-one bipartite matching between predicted b-boxes and ground-truth b-boxes using the Kuhn-Munkres algorithm~\cite{kuhn1955hungarian}. Let $\mathcal{M}' \subseteq B_p \times B_{gt}$ denote the optimal matching set that maximizes the total IoU:
\begin{equation}
\begin{aligned}
\mathcal{M} = &\arg\max_{\mathcal{M}' \subseteq B_p \times B_{gt}} \sum_{(i,j) \in \mathcal{M}'} \text{IoU}(b_i^p, b_j^{gt}),\\
&\text{s.t.} \quad \mathcal{M}' \text{ is one-to-one}.
\end{aligned}
\end{equation}

The pairwise IoU($PIoU$) score is then computed as the average over the matched pairs:

\begin{equation}
PIoU = 
\frac{1}{|\mathcal{M}|} \sum_{(i,j)\in \mathcal{M}} \frac{|b_i^p \cap b_i^{gt}|}{|b_i^p \cup b_i^{gt}|}.
\end{equation}

The final $HIoU$ is computed as the average of the global and pairwise IoU accuracy.
\begin{equation}
HIoU = \frac{GIoU + PIoU}{2}.
\end{equation}

\input{Figs/hiou}
\paragraph{Reasoning-format reward ($r_\text{format}$).} This reward encourages the model to generate reasoning outputs structured with the designated special token, specifically adhering to the format: \texttt{<think><area> ... </area><text> ... </text></think><answer> ... </answer>}. The \texttt{<area>...</area>} must contain a JSON-formatted representation of bounding boxes with depth, while \texttt{<text>...</text>} provides an explanation grounded in the specified spatial region. A reward of 1.0 is assigned to responses that strictly comply with this format.

\paragraph{Progressive-answer-accuracy reward ($r_{\text{ans},t}$).} This reward integrates both the correctness of the final answer and the progression of answer quality over time, yielding a more robust signal than purely rule-based evaluations. Specifically, we leverage an external Vision-Language Model (\textit{Doubao-1.5-vision-pro}) as the judge to assess response quality. The reward is defined as:
\begin{equation}
r_{\text{ans},t} = s_t + (s_t - \text{mean}\{s_{1,t-1}, \ldots, s_{N,t-1}\})
\end{equation}
where $s_t$ denotes the continuous score assigned by the Doubao judge at step $t$, based on the question, the predicted answer, and the ground-truth answer. The term $(s_t - \text{mean}\{s_{1,t-1}, \ldots, s_{N,t-1}\})$ captures the improvement between consecutive steps, thereby encouraging progressive refinement of the model's responses. The evaluation prompt used by the judger is provided in the Supp. Mat..

\paragraph{Correction-enhanced grounding reward ($r_{\text{bbox}}$).}
Given the structured nature of our output format, we can explicitly extract the sequence of bounding boxes generated during the reasoning process, thereby enabling fine-grained tracking of the step-by-step grounding. Let $B_{\text{ini}}$ denote the first bounding boxes in the reasoning trajectory that does not cover the entire image (i.e., $[0, 0, 1, 1]$), and let $B_{\text{end}}$ denote the final bounding boxes in the sequence. To assess the accuracy of the grounding results, we compute two intermediate metrics: (1) the $HIoU$ between the final box $B_{\text{end}}$ and the ground-truth box $B_{\text{gt}}$, denoted as $s_{\text{end}}=HIoU(B_{\text{end}},B_{gt})$; and (2) the $HIoU$ between the initial box $B_{\text{ini}}$ and $B_{\text{gt}}$, denoted as $s_{\text{init}}=HIoU(B_{\text{ini}},B_{gt})$. The overall grounding reward consists of two components: the final grounding accuracy $s_{\text{end}}$ and a correction-aware improvement term, defined as the relative gain $s_{\text{end}}-s_{\text{init}}$:
\begin{equation}
r_\text{bbox} = s_{\text{end}} + (s_{\text{end}}-s_{\text{init}}).
\end{equation}

\paragraph{Depth-consistency reward ($r_\text{depth}$).}
A spatially-aware model is expected to accurately capture the depth value associated with each specified region. However, the presence of hallucinations may cause the autoregressively generated next-token to become inconsistent with the actual input depth image, resulting in spatial confusion. To address this, we perform step-wise verification on the depth tokens generated during the reasoning process. Specifically, for each b-boxes-depth pair $(B_i, d_i)$, we extract the corresponding ground-truth depth $d_i^{gt}$ from the depth map $D_I$ based on $B_i$, and require the absolute error to be less than the threshold $\mathcal{T} = 0.1$. A reward is assigned only when all depth values throughout the reasoning trajectory meet this consistency criterion.
\begin{equation}
r_{\text{depth}} = \mathbb{I} \left(\forall i: \frac{|d_i - d_i^{\text{gt}}|}{d_i^{\text{gt}}} \leq \mathcal{T} \right),
\end{equation}
where $\mathbb{I}(\cdot)$ is the indicator function.

%% file: Sec/algo.tex
\begin{algorithm}[]
        \relsize{-1.5}
	\caption{Reverse Expansion for CoT Completion}
	\label{alg:reverse-expansion}

        \KwIn{One sample from the source dataset including question $Q$, image $I$, normalized b-boxes $B_{gt}$, and response $R$. Doubao-1.5-vision-pro model $\mathcal{M}_d$. DepthAnythingV2 model $\mathcal{M}_{DA}$.}
	\KwOut{Completed reasoning chain $R_{cot}$.}
	$B_0 \gets B_{gt}; \mathcal{B} \gets \{B_0\}; \mathit{ET} \gets False$; $K = 5$; $\mathcal{T} = 0.2$
    
	\tcp{Iteratively expand and merge b-boxes}
	\For{$t \gets 1$ \KwTo $K$}{
		$S \gets K - t + 1; B_t \gets \emptyset$

		\ForEach{b-box $b = (x_1,y_1,x_2,y_2) \in B_{t-1}$}{
			$\delta_{x_1} \gets \frac{x_1}{S}, \delta_{x_2} \gets \frac{1 - x_2}{S}, \delta_{y_1} \gets \frac{y_1}{S}, \delta_{y_2} \gets \frac{1 - y_2}{S}$
                
			$b' \gets (x_1 - \delta_{x_1}, y_1 - \delta_{y_1}, x_2 + \delta_{x_2}, y_2 + \delta_{y_2})$
            
                $B_t \gets B_t \cup \{b'\}$
		}
		
        
            \For{each pair $(b_i, b_j) \in B_t$ \textbf{where} $\text{IOU}(b_i,b_j) > 0$}
            {
                $b_{\text{merged}} \gets \big( \underset{k\in\{i,j\}}{\min} x_1^k,\ \underset{k\in\{i,j\}}{\min} y_1^k,\ \underset{k\in\{i,j\}}{\max} x_2^k,\ \underset{k\in\{i,j\}}{\max} y_2^k \big)$
            
                $B_t \gets \Big( B_t \setminus \{b_i, b_j\} \Big) \cup \{ b_{\text{merged}} \}$
            }
            
            $\mathcal{B} \gets \mathcal{B} \cup \{B_t\}$
            
            \tcp{Early terminate}
             \If{$|B_t| = 1$ \textbf{and not} $\mathit{ET}$}{$K \gets t+2$, $\mathit{ET} \gets True$}
            }
        \For{$i \gets 1$ \KwTo $\text{random choice from } \{0,1,2\}$}{
        \tcp{$\mathcal{U}_I(\mathcal{B})$:A set of b-boxes outside $\mathcal{B}$; $\mathcal{R_\mathcal{T}}(\cdot)$: Random selection with an area difference no more than $\mathcal{T}$}
    	$B_{K+i} = \mathcal{R_\mathcal{T}}(\mathcal{U}_I(\bigcup_{j=K}^{K+i-1} B_j)\bigr); \mathcal{B} \gets \mathcal{B} \cup \{B_{K+i}\}$
        }
	\tcp{Reverse b-boxes set to generate CoT}
	$\mathcal{B}_{rev} \gets \text{reverse}(\mathcal{B});   I_{seq} \gets \text{DrawBBoxes}(I, \mathcal{B}_{rev})$

        $D_{I} \gets \mathcal{M}_{DA}(I); R_{cot} \gets \mathcal{M}_d\bigl(Q,\: R,\: D_{I},\: (I_{seq},\: \mathcal{B}_{rev})\bigr)$
	
	\Return $R_{cot}$
\end{algorithm}

%% file: Figs/hiou.tex
\begin{figure}[!htbp]
    \centering
    \vspace{-1.2em}
    \includegraphics[width=0.46\textwidth]{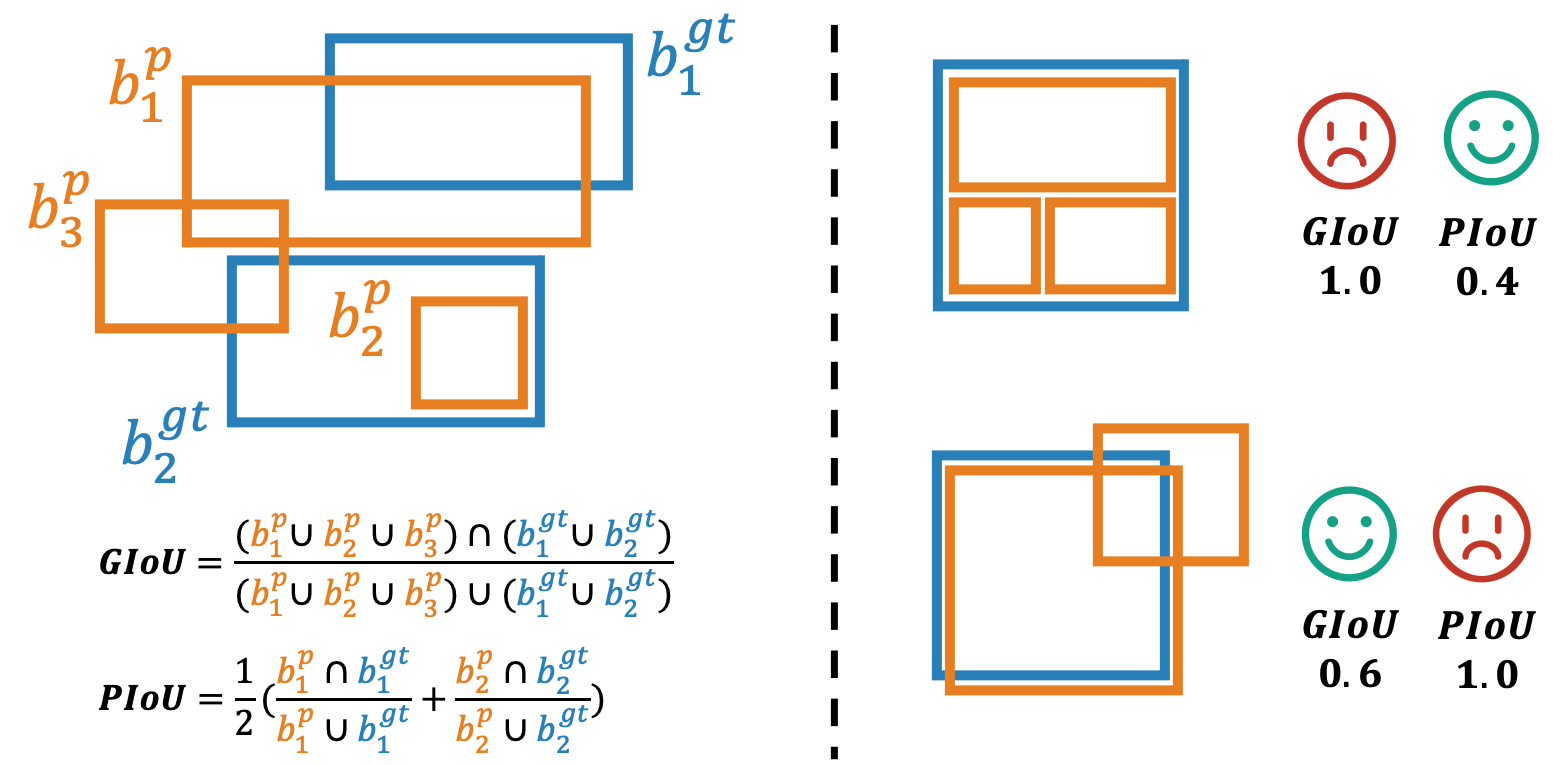}
    \caption{Visualization of our proposed $HIoU$ (left). 
    The performance of $GIoU$ and $PIoU$ are illustrated respectively (right), highlighting the robustness against reward hacking.}\label{fig:hiou}
    \vspace{-0.8em}
\end{figure}

%% file: Sec/exp.tex
\section{Experiments}\label{sec:exp}
We evaluate \method with several state-of-the-art (SOTA) methods
on an array of different categories as follows. More details about datasets and the evaluation metric are listed in Supp. Mat..

\subsection{Spatial Intelligence}
\input{Tabs/tab_spatial}
We evaluate our method against several SOTA methods on a range of spatial understanding benchmarks. Benefiting from our spatially-aware think-with-images training paradigm, our model demonstrates superior 3D understanding capabilities. As shown in Tab.~\ref{table:spatial}, under the same base model, our method outperforms SpatialBot~\cite{cai2024spatialbot} by \textbf{7.82\%} (64.3 vs. 59.6) and SSR~\cite{liu2025ssr} by \textbf{11.17\%} (74.5 vs. 67.1) on SpatialBench~\cite{cai2024spatialbot}. Furthermore, we evaluate our method on larger-scale benchmarks, SAT (Static)~\cite{ray2024sat} and CV-Bench~\cite{cvbench}, achieving gains of \textbf{11.15\%} (72.8 vs. 65.5) and \textbf{3.97\%} (75.9 vs. 73.0) over the Qwen2.5-VL-7B base model, respectively. Although both SpatialBot and SSR incorporate depth images to enhance spatial understanding, \emph{we argue that depth perception and spatial grounding are inherently complementary}. By introducing reasoning over spatially grounded regions, our method achieves more significant improvements.

We further conducted a comparison with the representative SOTA closed-source model—ChatGPT-o3~\cite{openai2025o3o4mini}. On SpatialBench, \method achieves a comparable average score to o3 (74.6 vs. 74.8). Notably, on SAT-Static, our method even outperforms o3 by a significant margin of \textbf{8.01\%} (72.8 vs. 67.4), demonstrating the superior capability of \method in spatial perception.

\input{Tabs/tab_visual}
\input{Tabs/tab_grounding}
\subsection{Visual Perception}
In this section, we comprehensively evaluate the visual perception capabilities of the method in terms of visual understanding, grounding capability and self-correction ability.
\paragraph{Visual Understanding.} We select scene-related (e.g. non-planar) subsets from the VisCoT as VisCoT\_s, and choose the attribute and spatial subsets from V*Bench. As shown in Tab.~\ref{table:visual}, on VisCoT\_s dataset, under the same base model -- LLaVA-1.5-7B, \method outperforms VisCoT by \textbf{11.76\%} (0.751 vs. 0.672). Taking Qwen2.5-VL-7B as the base model for training, we surpass VisRL by \textbf{8.89\%} (0.760 vs. 0.698). V*Bench presents a more challenging evaluation of fine-grained perception on high-resolution images. Remarkably, our method outperforms the state-of-the-art method -- SEAL by \textbf{5.75\%} on the Attribute subset (0.791 vs. 0.748), and by \textbf{1.70\%} on the Spatial subset (0.776 vs. 0.763). In contrast to VisCoT, VisRL and SEAL, \emph{\method does not rely on a staged crop-image process}. 

\input{Figs/comparison}
ChatGPT-o3 leverages an external image magnification tool to explicitly capture fine-grained regions within the image. In contrast, \method adopts an intrinsic integration of visual information without any external tools. As shown in Tab.~\ref{table:visual}, a detailed comparison reveals that our method achieves comparable overall performance to o3 on the VisCoT\_s benchmarks (0.760 vs. 0.762). Furthermore, on V*Bench-Attribute, our method surpasses o3 by a substantial margin of \textbf{7.04\%} (0.791 vs. 0.739), highlighting the effectiveness in fine-grained visual reasoning. Instead, \method integrates visual perception directly into the thinking process with dynamically evolving bounding boxes to represent shifts in attention. This results in a more unified and intrinsically coherent reasoning approach. Furthermore, we explicitly incorporate 3D spatial information, which brings additional benefits to scene-level grounding.

\paragraph{Grounding Capability.} Since the accuracy of the final response is partly dependent on the correctness of the bounding boxes generated during the think process, we further evaluate the model’s grounding capability as shown in Tab.~\ref{table:grounding}. Specifically, we select two structurally similar tasks—Referring Expression Comprehension (REC) and Open-Vocabulary Detection (OVD)—both of which require the model to generate bounding boxes conditioned on textual descriptions. \method outperforms all previous generalist models with comparable parameters, achieving an average improvement of \textbf{1\%} to \textbf{3\%} over VisRL (93.08 vs. 91.78 on RefCOCO~\cite{refcoco}, 85.76 vs. 83.90 on RefCOCO+~\cite{refcoco+/g}, 90.47 vs. 88.82 on RefCOCOg~\cite{refcoco+/g}).  Moreover, in most of cases, \method even surpasses previous state-of-the-art specialist models (e.g. Grounding-DINO, UNINEXT). To further assess multi-object grounding performance of our method, we adopt OVDEval~\cite{ovdeval} with NMS-AP as the evaluation metric. Empowered by our proposed $HIoU$, \method exhibits promising robustness in multi-object scenarios, outperforming VisRL by \textbf{45.9\%} (37.8 vs. 25.9). 

\input{Tabs/tab_general}
\input{Figs/visual}

\paragraph{Self-correction Ability.} \emph{An effective image focus mechanism can simultaneously enable accurate attention correction.} To this end, we also highlight the capability for bounding box correction shown in Fig.~\ref{fig:visual}: given a starting focused area, \method is able to judge, search, and ultimately attend to the prompt-relevant area. Furthermore, we compare the visualized results of \method (\underline{red} b-boxes) with VisRL (\underline{green} b-boxes) and VLM-R1 (\underline{blue} b-boxes) on the OVD task. \method successfully detects multiple relevant objects, while both VisRL and VLM-R1 yield only single-object predictions. This further highlights the multi-object friendliness of \method.

Overall, as shown in Fig.~\ref{fig:comparison}, at the 7B parameter scale, \method (\underline{orange}) exhibits strong reasoning capabilities in both spatial understanding and visual perception, outperforming all open-source models (\underline{blue}) and approaching the performance of the larger-scale ChatGPT-o3 (\underline{green}). Notably, \method even surpasses ChatGPT-o3 on V*Bench and SAT-Static.
\subsection{General VLM Benchmarks}
As shown in Tab.~\ref{table:general}, we report results on widely used general benchmarks, including MME~\cite{mme} perception (MME\textsuperscript{P}), MME cognition (MME\textsuperscript{C}), MMBench~\cite{mmb} test and dev sets (denoted as MMB\textsuperscript{T} and MMB\textsuperscript{D}), SEED-Bench~\cite{seedbench} images (SEED-I), VQA\textsuperscript{v2}~\cite{vqa} test-dev split, and POPE~\cite{pope} (measured as the average F1 score over three categories on the COCO validation set). Across most of these benchmarks, \method not only avoids performance degradation but even achieves notable improvements, demonstrating the robustness of our method—particularly in scenarios that depth information can benefit. Under the same base models, \method consistently outperforms both VisCoT, which focuses on fine-grained visual perception, and SpatialBot, which emphasizes spatial reasoning. Remarkably, on MMB\textsuperscript{T}, \method achieves about \textbf{4\%} improvements across different base model settings (69.3 vs. 66.5 on LLaVA-1.5-7B, 76.8 vs. 73.7 on Bunny-Llama3-8B, 83.4 vs. 80.3 on Qwen2.5-VL-7B ).

\subsection{Ablation Study}
\input{Tabs/tab_ablation}
\paragraph{Training Strategy.} We present a comprehensive ablation study in Tab.~\ref{table:ablation}.
VQA-SFT refers to directly applying SFT on the original question-answer pairs, which serve as the source data from where \textbf{SIF-50K} was constructed, whereas CoT-SFT leverages the Chain-of-Thought (CoT) construction strategy introduced in Alg.~\ref{alg:reverse-expansion}. This demonstrates that guiding the model to think with images yields a notable performance improvement of \textbf{8.58\%} (0.582 vs. 0.536). However, SFT alone mainly helps the model learn output formatting, and in some cases (e.g., GQA), even leads to performance degradation. In contrast, incorporating RL yields consistent and substantial improvements, achieving an additional \textbf{30.58\%} gain over SFT alone (0.760 vs. 0.582). 
We further perform ablations on various RL rewards (w/o $r_{\text{ans},t}, r_{\text{bbox}}, r_{\text{depth}}$) and evaluate the impact of depth information (w/o $D_{I}$). The results suggest that the observed performance gains are primarily attributed to three key factors:
1) the think-with-images reasoning paradigm, which promotes spatially grounded cognition;
2) the carefully designed reward functions for both bounding box prediction and response generation, which work synergistically to encourage iterative correction and refinement;
3) the inclusion of depth inputs, which enhance the model’s spatial intelligence during grounding. Together, these designs form a unified and robust framework for spatially-aware visual grounding, endowing the model with a general-purpose reasoning capability that improves performance across diverse benchmarks.

\paragraph{Component Removal.} We further perform ablations on various RL rewards (w/o $r_{\text{ans},t}, r_{\text{bbox}}, r_{\text{depth}}$) and evaluate the impact of depth information (w/o $D_{I}$) in Tab.~\ref{table:ablation}. The results suggest that the observed performance gains are primarily attributed to three key factors:
1) the think-with-images reasoning paradigm, which promotes spatially grounded cognition;
2) the carefully designed reward functions for both bounding box prediction and response generation, which work synergistically to encourage iterative correction and refinement;
3) the inclusion of depth inputs, which enhance the model’s spatial intelligence during grounding. Together, these designs form a unified and robust framework for spatially-aware visual grounding, endowing the model with a general-purpose reasoning capability that improves performance across diverse benchmarks.

\paragraph{Effectiveness with RGB-Depth Input.} When both RGB and depth are provided, \method still outperforms baselines—e.g., Qwen2.5-VL: ours vs. VisRL = \textbf{0.760} vs. 0.691; LLaVA-1.5: ours vs. VisCoT = \textbf{0.751} vs. 0.674. 

\paragraph{Ablation on Correction-enhanced Grounding Reward.} Removing ($s_{end} - s_{init}$) reduces performance on VisCoT\_s from \textbf{0.760} to 0.723. We also experimented with a step-wise non-decreasing-IoU reward, which also drops from \textbf{0.760} to 0.735 due to restricted exploration.

%% file: Tabs/tab_spatial.tex
\begin{table}[t!]
    \centering
    \footnotesize
    \setlength{\tabcolsep}{1mm}
    \begin{tabular}{lcccccc}
    \hline
                                                 & \multicolumn{4}{c}{\textbf{SpatialBench}}                                                                                          &                                       &                                     \\
    \multirow{-2}{*}{\textbf{Model}}             & Pos.                     & Ext.                    & Cnt.                     & Size                                 & \multirow{-2}{*}{\textbf{SAT-Sta.}} & \multirow{-2}{*}{\textbf{CV-Bench}} \\ \hline
    o3-2025-04-16                                 & 79.4                         & 95.0                         & 89.9                         & 35.0                                  & 67.4                                     & —                              \\ \hline
    LLaVA-1.5-7B                                 & 44.1                         & 45.0                         & 82.8                         & 30.0                                  & 49.8                                     & 51.7                                \\
    LLaVA-NeXT-7B                                & 47.1                         & 75.0                         & 84.0                         & 20.0                                  & 54.1                                     & 62.7                                \\
    SpatialBot-3B                                & 50.0                         & 80.0                         & 86.7                         & 25.0                                  & 61.5                                     & 65.1                                   \\
    Emu3-8B                                      & 47.1                         & 20.0                         & 10.0                         & 25.0                                  & —                                     & —                                   \\ \hline
    \cellcolor[HTML]{FFFFFF}Bunny-8B      & \cellcolor[HTML]{FFFFFF}50.0 & \cellcolor[HTML]{FFFFFF}75.0 & \cellcolor[HTML]{FFFFFF}90.4 & \cellcolor[HTML]{FFFFFF}\textbf{25.0} & 60.8                                     & 61.0                                   \\
    \cellcolor[HTML]{FFFFFF}SpatialBot-8B & \cellcolor[HTML]{FFFFFF}53.0 & \cellcolor[HTML]{FFFFFF}75.0 & \cellcolor[HTML]{FFFFFF}90.4 & \cellcolor[HTML]{FFFFFF}20.0          & —                                     & —                                   \\
    \rowcolor[HTML]{E7E6E6} 
    \method-8B              & \textbf{61.8}                & \textbf{80.0}                & \textbf{92.2}                & 23.3                                  & \textbf{67.9}                                    & \textbf{65.6}                                  \\ \hline
    Qwen2.5-VL-7B                                & 61.8                         & 80.0                         & 87.1                         & 30.0                                  & 65.5                                  & 73.0                                \\
    \cellcolor[HTML]{FFFFFF}SSR-7B               & \cellcolor[HTML]{FFFFFF}64.7 & \cellcolor[HTML]{FFFFFF}85.0 & \cellcolor[HTML]{FFFFFF}90.2 & \cellcolor[HTML]{FFFFFF}28.3          & —                                     & 73.3                                \\
    \rowcolor[HTML]{E7E6E6} 
    \method-7B               & \textbf{73.5}                & \textbf{95.0}                & \textbf{94.7}                & \textbf{35.0}                         & \textbf{72.8}                         & \textbf{75.9}                       \\ \hline
    \end{tabular}
    \vspace{-0.8em}
    \caption{\textbf{Spatial perception evaluation} on SpatialBench (Position, Existence, Counting, Size), SAT (Static), and CV-Bench. Bunny-LLaMA3-8B and Qwen2.5-VL-7B serve as base models in the 3-rd and 4-th groups, respectively. The \textbf{best} is highlighted. }\label{table:spatial}
    \vspace{-1.8em}
\end{table}

%% file: Tabs/tab_visual.tex
\begin{table*}[t!]
    \centering
    \footnotesize
    \setlength{\tabcolsep}{1.7mm}
    \begin{tabular}{lcccccccc}
    \hline
                                      & \multicolumn{6}{c}{\textbf{VisCoT\_s}}                                                                                                                           & \multicolumn{2}{c}{\textbf{V*Bench}}           \\
    \multirow{-2}{*}{\textbf{Model}}  & Flickr30k                     & GQA                           & Open images                   & VSR                           & CUB    & Avg                       & Attribute                     & Spatial        \\ \hline
    o3-2025-04-16~\cite{openai2025o3o4mini}                      & \cellcolor[HTML]{FFFFFF}0.828 & 0.706                       & 0.515                         & 0.826                         & 0.933   & 0.762                      & \cellcolor[HTML]{FFFFFF}0.739 & 0.803         \\ \hline
    LLaVA-1.5-7B~\cite{mllm:liu2023improvedllava}                      & \cellcolor[HTML]{FFFFFF}0.581 & 0.534                         & 0.412                         & 0.572                         & 0.530   & 0.526                       & \cellcolor[HTML]{FFFFFF}0.435 & 0.566         \\
    VisCoT-LLaVA-1.5-7B~\cite{shao2024visual}                         & 0.671                         & 0.616                         & \textbf{0.833}                & 0.682                         & 0.556    & 0.672                     & 0.466                             & 0.571             \\
    \rowcolor[HTML]{E7E6E6} 
    \method-LLaVA-1.5-7B & \textbf{0.749}                & \textbf{0.675}                & 0.801                         & \textbf{0.698}                & \textbf{0.831}     & \textbf{0.751}           & \textbf{0.565}                & \textbf{0.671} \\ \hline
    Qwen2.5-VL-7B~\cite{bai2025qwen2.5}                     & \cellcolor[HTML]{FFFFFF}0.601 & 0.467                         & 0.289                         & 0.581                         & 0.583                        & 0.504 & \cellcolor[HTML]{FFFFFF}0.644 & 0.634          \\
    \cellcolor[HTML]{FFFFFF}VisRL-Qwen2.5-VL-7B\#~\cite{chen2025visrl}    & \cellcolor[HTML]{FFFFFF}0.662 & \cellcolor[HTML]{FFFFFF}0.589 & \cellcolor[HTML]{FFFFFF}0.767 & \cellcolor[HTML]{FFFFFF}0.698 & 0.772 & \cellcolor[HTML]{FFFFFF}0.698 & \cellcolor[HTML]{FFFFFF}0.678 & 0.658          \\
    \cellcolor[HTML]{FFFFFF}SEAL-7B (V*)~\cite{rela:vstar}    & 0.723                            & 0.599                             & 0.448                             & 0.730                             & 0.640                             & 0.628 & \cellcolor[HTML]{FFFFFF}0.748 & 0.763          \\
    \rowcolor[HTML]{E7E6E6} 
    \method-Qwen2.5-VL-7B  & \textbf{0.755}                & \textbf{0.664}                & \textbf{0.773}                & \textbf{0.734}                & \textbf{0.872}    & \textbf{0.760}            & \textbf{0.791}                & \textbf{0.776} \\ \hline
    \end{tabular}
    \vspace{-0.8em}
    \caption{\textbf{Visual perception performance} on VisCoT\_s and V*Bench. \# indicates methods trained on the same \textbf{SIF-50K} dataset as ours. For the same base models, the \textbf{best} is highlighted.}\label{table:visual}
    \vspace{-0.8em}
\end{table*}

%% file: Tabs/tab_grounding.tex
\begin{table*}[t!]
    \centering
    \footnotesize
    \begin{tabular}{lccccccccc}
    \hline
                                     & \multicolumn{3}{c}{\textbf{RefCOCO}}          & \multicolumn{3}{c}{\textbf{RefCOCO+}}         & \multicolumn{2}{c}{\textbf{RefCOCOg}} &                                                         \\
    \multirow{-2}{*}{\textbf{Model}} & val           & test-A        & test-B        & val           & test-A        & test-B        & val-u             & test-u            & \multirow{-2}{*}{\textbf{OVDEval}} \\ \hline
    UNIEXT~\cite{uninext}                           & 92.6          & 94.3          & \textbf{91.5} & 85.2          & 89.6          & 79.8          & 88.7              & 89.4              & —                                                       \\
    Grounding-DINO~\cite{groundingdino}                   & 90.6          & 93.2          & 88.2          & 82.8          & 89.0          & 75.9          & 86.1              & 87.0              & 25.3                                                    \\ \hline
    Qwen2.5-VL-3B~\cite{bai2025qwen2.5}                    & 89.1          & 91.7          & 84.0          & 82.4          & 88.0          & 74.1          & 85.2              & 85.7              & 25.5                                                    \\
    Qwen2.5-VL-7B~\cite{bai2025qwen2.5}                     & 90.0          & 92.5          & 85.4          & 84.2          & 89.1          & 76.9          & 87.2              & 87.2              & 29.1                                                    \\
    VLM-R1-Qwen2.5-VL-3B~\cite{shen2025vlmr1}             & 91.2          & 92.9          & 87.3          & 84.8          & 88.1          & 76.8          & 87.8              & 87.9              & 31.0                                                    \\
    VLM-R1-Qwen2.5-VL-7B\#~\cite{shen2025vlmr1}            & 90.9          & 93.4          & 87.5          & 85.0          & 88.6          & 77.2          & 88.2              & 88.3              & 26.3                                                    \\
    VisRL-Qwen2.5-VL-7B\#~\cite{chen2025visrl}                           & 92.3          & 94.2          & 88.9          & 84.4          & 89.5          & 77.8          & 88.5              & 89.1              & 25.9                                                    \\
    \rowcolor[HTML]{E7E6E6} 
    \method-Qwen2.5-VL-7B               & \textbf{93.8} & \textbf{95.1} & 90.4          & \textbf{86.0} & \textbf{90.7} & \textbf{80.5} & \textbf{90.4}     & \textbf{90.6}     & \textbf{37.8}                                           \\ \hline
    \end{tabular}
    \vspace{-0.8em}
    \caption{Performance (Top-1 Accuracy@0.5) on Referring Expression Comprehension tasks and performance (NMS-AP) on Open-Vocabulary Detection tasks. \# indicates methods trained on the same \textbf{SIF-50K} dataset as ours. The \textbf{best} is highlighted. }\label{table:grounding}
    \vspace{-1.5em}
\end{table*}

%% file: Figs/comparison.tex
\begin{figure}[]
    \centering
    \includegraphics[width=0.48\textwidth]{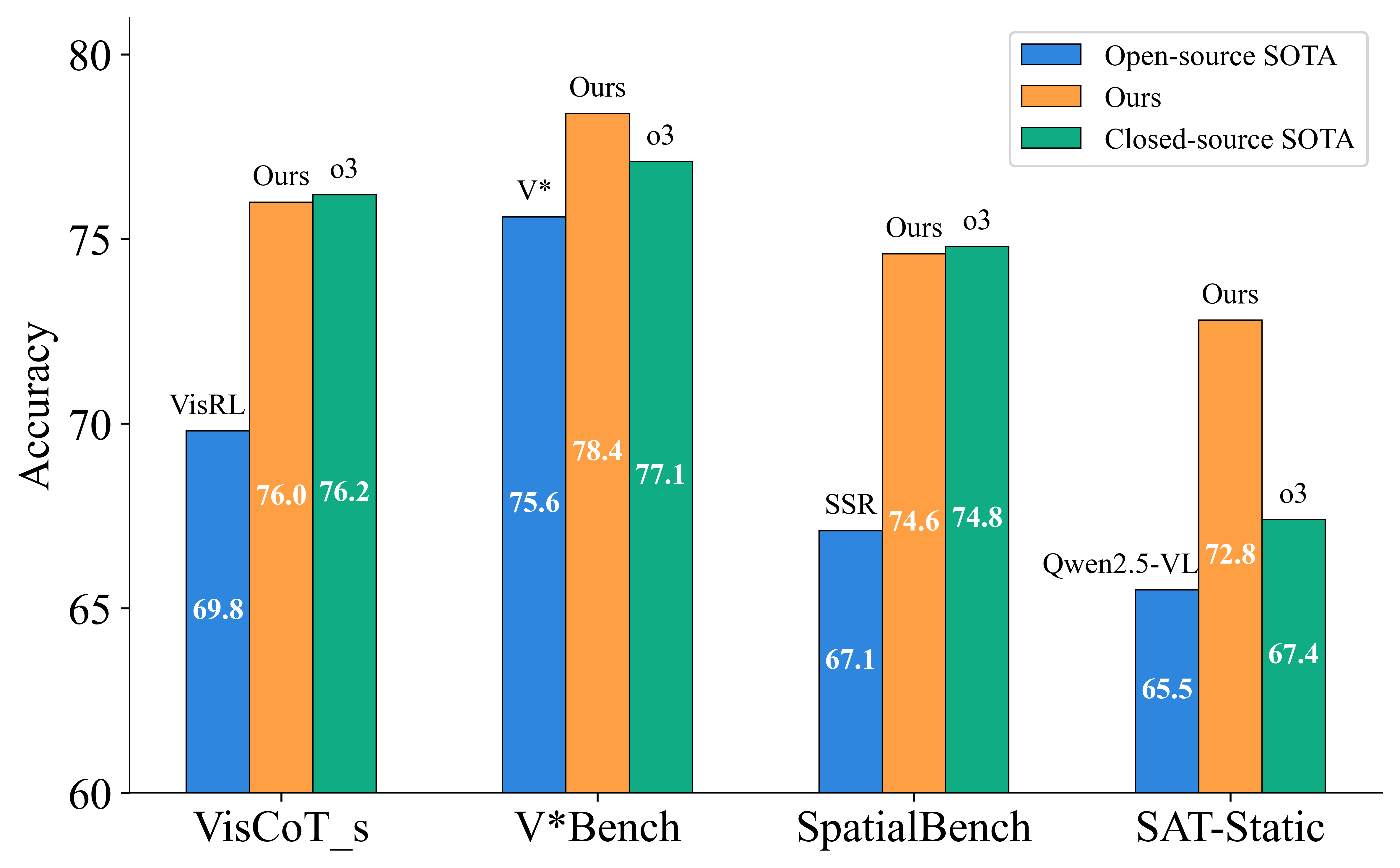}
    \vspace{-0.8em}
    \caption{Comparison with  open-source SOTA methods (\underline{blue}) under various benchmarks in terms of the same base model. Besides, we also include the performance of the proprietary SOTA model ChatGPT-o3-2025-04-16 (\underline{green}).}\label{fig:comparison}
    \vspace{-0.8em}
\end{figure}

%% file: Tabs/tab_general.tex
\begin{table}[]
    \centering
    \footnotesize
    \setlength{\tabcolsep}{0.2mm}
    \begin{tabular}{lccccccc}
    \hline
    \textbf{Model}                & \textbf{MME\textsuperscript{P}} & \textbf{MME\textsuperscript{C}} & \textbf{MMB\textsuperscript{T}} & \textbf{MMB\textsuperscript{D}} & \textbf{SEED-I} & \textbf{VQA\textsuperscript{v2}} & \textbf{POPE} \\ \hline
    LLaVA               & 1511            & 282             & 66.5            & 62.1            & 65.8            & 79.1             & 85.9          \\
    VisCoT                  & 1454            & \textbf{308}    & 69.2            & \textbf{66.6}   & 66.0            & 81.3             & 86.0          \\
    \rowcolor[HTML]{E7E6E6} 
    \method  & \textbf{1531}   & 295             & \textbf{69.3}   & 62.8            & \textbf{66.0}   & \textbf{81.6}    & \textbf{86.6} \\ \hline
    Bunny             & 1574            & 342             & 73.7            & 74.2            & 72.3            & 80.5             & 85.2          \\
    SpatialBot       & 1576            & 333             & 75.8            & 74.8            & 72.4            & 80.9             & 85.3          \\
    \rowcolor[HTML]{E7E6E6} 
    \method & \textbf{1580}   & \textbf{355}    & \textbf{76.8}   & \textbf{75.0}   & \textbf{72.5}   & \textbf{81.8}    & \textbf{85.7} \\ \hline
    Qwen2.5VL               & 1670            & 623             & 80.3            & 81.5            & 77.0            & 83.3             & 85.9          \\
    \rowcolor[HTML]{E7E6E6} 
    \method  & \textbf{1702}   & \textbf{650}    & \textbf{83.4}   & \textbf{82.4}   & \textbf{77.2}   & \textbf{84.5}    & \textbf{86.9} \\ \hline
    \end{tabular}
    \vspace{-0.8em}
    \caption{Results on general VLM Benchmarks. Methods are grouped by their underlying base model: LLaVA-1.5-7B (top), Bunny-Llama3-8B (middle), and Qwen2.5-VL-7B (bottom). For each group, the \textbf{best} is highlighted. }\label{table:general}
    \vspace{-1.2em}
\end{table}

%% file: Figs/visual.tex
\begin{figure}[!htbp]
    \centering
    \includegraphics[width=0.48\textwidth]{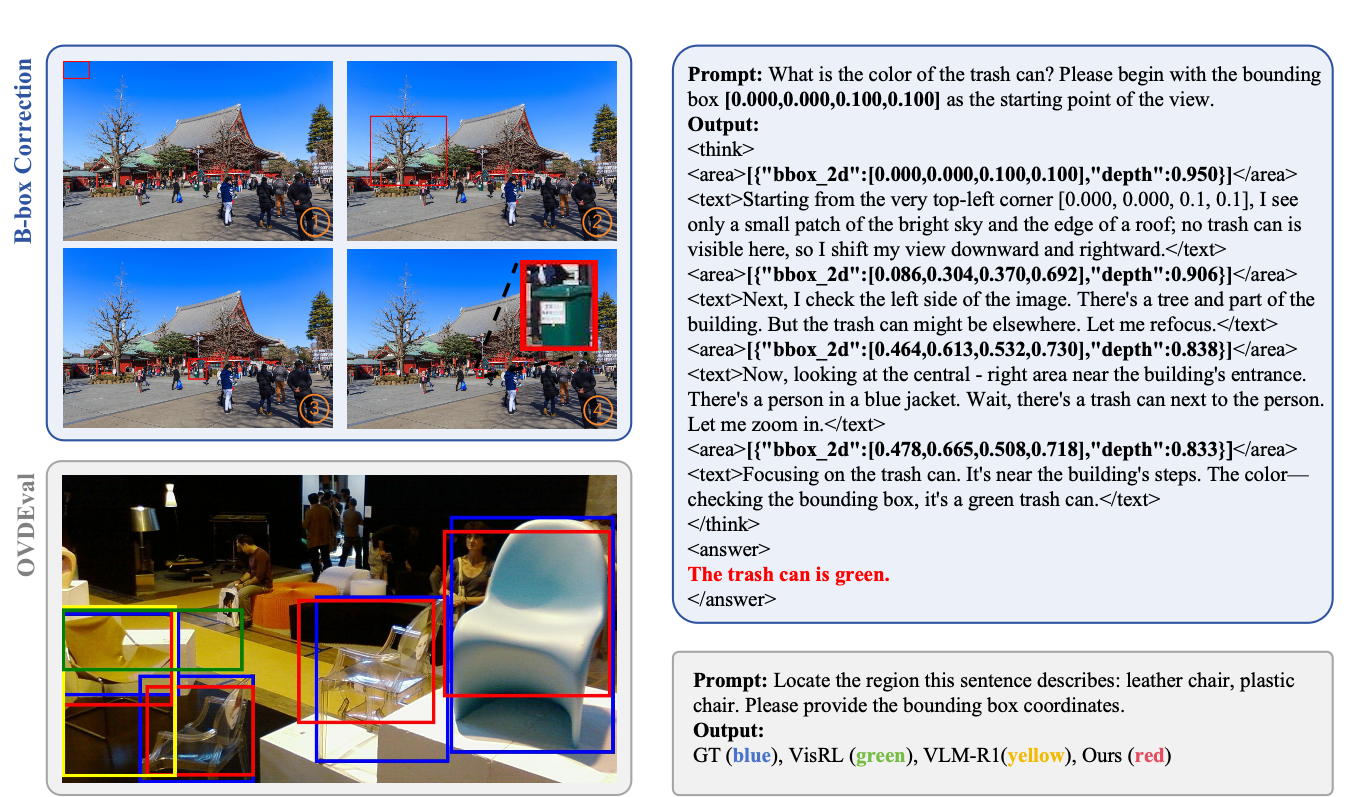}
    \vspace{-0.8em}
    \caption{Visualization of the \method’s region correction and detection (multi-objects) capabilities.}\label{fig:visual}
    \vspace{-1.8em}
\end{figure}

%% file: Tabs/tab_ablation.tex
\begin{table}[t!]
    \centering
    \footnotesize
    \setlength{\tabcolsep}{0.8mm}
    \begin{tabular}{lcccccc}
    \hline
                                     & \multicolumn{5}{c}{\textbf{VisCoT\_s}}                                                & \multicolumn{1}{l}{} \\
    \multirow{-2}{*}{\textbf{Model}} & Flickr30k      & GQA            & Open images    & VSR            & CUB            & Avg                  \\ \hline
    \rowcolor[HTML]{FFFFFF} 
    VQA-SFT                          & 0.542          & 0.508          & 0.390          & 0.597          & 0.642          & 0.536               \\
    \rowcolor[HTML]{FFFFFF} 
    CoT-SFT                          & 0.517          & 0.468          & 0.556          & 0.588          & 0.780          & 0.582               \\
    \rowcolor[HTML]{FFFFFF} 
    w/o $r_\text{bbox}$                  & 0.641          & 0.580          & 0.711          & 0.705          & 0.839          & 0.695               \\
    \rowcolor[HTML]{FFFFFF} 
    w/o $r_{\text{ans},t}$              & 0.628          & 0.573          & 0.658          & 0.663          & 0.795          & 0.663               \\
    \rowcolor[HTML]{FFFFFF} 
    w/o $D_{I}$                & 0.571          & 0.595          & 0.589          & 0.651          & 0.870          & 0.655               \\
    \rowcolor[HTML]{FFFFFF} 
    w/o $r_\text{depth}$                 & 0.748          & \textbf{0.666} & 0.762          & 0.718          & \textbf{0.872} & 0.753                \\
    \rowcolor[HTML]{E7E6E6} 
    Full   & \textbf{0.755} & 0.664          & \textbf{0.773} & \textbf{0.734} & \textbf{0.872} & \textbf{0.760}       \\ \hline
    \end{tabular}
    \vspace{-0.8em}
    \caption{Performance on different ablated settings in terms of Qwen2.5-VL-7B.}\label{table:ablation}
    \vspace{-1.8em}
\end{table}

%% file: Sec/conclusion.tex
\section{Conclusion and Limitation}\label{sec:conclusion}
In this paper, we propose \method, a spatially-aware image-text interleaved reasoning framework. Inspired by human-like prompt-driven search in 3D environments, \method performs spatially-aware grounding before delivering the final response. Specifically, we introduce a novel pipeline for generating CoT datasets tailored for think-with-images reasoning, enabling process-level supervision. Building on this dataset, we propose GRPO-SIF by incorporating not only region-level corrective signals, but also proposing progress learning and depth consistency rewards. Extensive experiments across diverse benchmarks demonstrate the effectiveness of \method.

\paragraph{Limitation \& Future Work.}
Since \method is trained on single-image, it may face challenges in generalizing to dynamic spatial scenarios that require reasoning across multiple images. We believe extending it to such settings would be interesting for higher practical impact.

\section{Acknowledgments} 
This work was supported by the National Natural Science Foundation of China under contract No. 62171256.
 

%% file: Sec/supmat.tex
\label{sec:appendix}
In this supplementary material, we provide more technical details and experimental results, including 1) the overview of our training dataset; 2) the benchmarks and evaluation metric used in this paper; 3) some implementation details in terms of experimental setup, hyper-parameters setup and prompt designs; and 4) more visualization of \method.

\input{Tabs/tab_dataset}

\section{Training Dataset}\label{sec:training dataset}
We propose \textbf{SIF-50K} dataset for training based on our CoT data generation pipeline. The data are from VisCoT training set~\cite{shao2024visual} and TallyQA~\cite{acharya2019tallyqa}, with each source sample containing a question, image, answer, and ground-truth bounding boxes. Dataset statistics are provided in Tab.~\ref{table: dataset}.

\section{Benchmarks}\label{sec:benchmarks}
We conduct evaluations on a series of benchmarks, following the metric settings defined by each benchmark.
\subsection{Spatial Intelligence}
\paragraph{SpatialBench:} We evaluate the spatial comprehension capabilities of MLLMs using SpatialBench~\cite{cai2024spatialbot}, which contains manually annotated question-answer pairs focused on spatial understanding and reasoning. We use four categories: position (34 samples), existence (40 samples), counting (20 samples), and size (40 samples).
\paragraph{SAT (Static):} The SAT~\cite{ray2024sat} dataset includes both static and dynamic spatial reasoning tasks. We select the static evaluation subset (single-image), which contains 127405 samples.
\paragraph{CV-Bench:} With 2,638 manually inspected samples, CV-Bench~\cite{cvbench} includes four tasks: spatial relationship, object counting, depth order, and relative distance.
\subsection{Visual Perception}
\paragraph{VisCoT\_s:} VisCoT\_s is the subset from VisCoT dataset (eval)~\cite{shao2024visual}, comprising several scene-specific datasets (3D information may help), including Flickr30k~\cite{flickr30k}, VSR~\cite{vsr}, GQA~\cite{gqa}, Open Images~\cite{openimages}, and CUB~\cite{cub}. Specifically, Flickr30k~\cite{flickr30k} dataset provides five human-written captions per image along with bounding box annotations for most referenced objects. Building upon this, ~\cite{shao2024visual} further utilizes GPT-4 to generate questions that specifically target small objects. The Visual Spatial Reasoning (VSR)~\cite{vsr}, GQA~\cite{gqa}, and Open Images~\cite{openimages} datasets are rich in spatial relational information between image entities. The Birds-200-2011 (CUB)~\cite{cub} dataset is a widely adopted benchmark for fine-grained visual categorization, which contains high-resolution bird images along with detailed part annotations, attribute labels, and bounding boxes. To better exploit this dataset in the context of MLLMs, ~\cite{shao2024visual} design probing questions that require the model to identify fine-grained bird characteristics, thereby assessing its capacity for detailed visual recognition.
\paragraph{V* Bench:} To evaluate the performace of MLLMs in complex visual scenarios with dense, high-resolution imagery, we use V*Bench~\cite{rela:vstar}, a benchmark of 191 images (avg. resolution: 2246×1582) comprising two tasks: attribute recognition (115 samples) and spatial relationship reasoning (76 samples). These tasks assess models’ fine-grained visual understanding.
\subsection{Visual Grounding}
\paragraph{RefCOCO/RefCOCO+/RefCOCOg:} Referring Expression Comprehension (REC) can directly use the Intersection-over-Union (IoU) between the predicted and ground-truth bounding boxes as the explicit evaluation. Thus, We evaluate various methods on REC benchmarks, including RefCOCO~\cite{refcoco}, RefCOCO+~\cite{refcoco+/g}, and RefCOCOg~\cite{refcoco+/g}. RefCOCO and RefCOCO+ were collected via an interactive game and follow the standard val/testA/testB split, where testA focuses on people and testB on other objects. RefCOCO+ excludes absolute spatial terms from queries. RefCOCOg, collected non-interactively, features longer and more descriptive queries. We follow ~\cite{shao2024visual} by setting the IoU threshold to 0.5 for accuracy evaluation, i.e., Top-1 Accuracy@0.5 as evaluation metric.
\paragraph{OVDEval:} OVDEval~\cite{ovdeval} is a benchmark for the Open-Vocabulary Detection task, comprising 9 sub-tasks that evaluate commonsense reasoning, attribute and spatial understanding, object relations, etc.. Compared to RefCOCO/RefCOCO+/RefCOCOg, OVDEval supports multi-object grounding with multiple bounding box annotations. Additionally, we adopt the Non-Maximum Suppression Average Precision (NMS-AP) metric from OVDEval~\cite{ovdeval} for evaluation.
\subsection{General VLM Benchmarks}
We further evaluate the models on five general benchmarks, as summarized in Tab. 4 of the main text: MME\cite{mme}, which assesses perception and cognition across 14 sub-tasks; MMBench\cite{mmb}, a systematically designed benchmark covering 20 capability dimensions for robust, holistic evaluation; SEED-Bench~\cite{seedbench}, containing 19242 multiple-choice questions with high-quality human annotations—spanning 12 evaluation dimensions across image and video modalities; VQA~\cite{vqa}, a dataset of open-ended questions over 265016 images (from COCO and abstract scenes), requiring vision, language, and commonsense understanding; and POPE~\cite{pope}, which frames hallucination evaluation as binary object presence questions. More details about evaluation splits have been detailed in Sec. 4 of the main text.

\section{Implementation details}\label{sec:implementation}
\subsection{Experimental Setup}\label{sec:env}
We perform all the experiments on a machine with 8*NVIDIA H20 96GB and Intel(R) Xeon(R) Platinum 8457C with 180 cores. 

\subsection{Hyper-parameters setup}
Taking Qwen2.5-VL-7B as an example, during the SFT stage, we utilized the full SIF-50k dataset, setting LORA rank (i.e., $r$) to 8, training for 3 epochs with a learning rate of 1e-4.
In the RL stage, we sampled only 200 samples from the SIF-50k dataset. We adopted the default GRPO hyperparameter settings, configuring $N$ to 8, the KL divergence ratio (i.e., $\beta$) to 0.04, LORA rank (i.e., $r$) to 64, and trained for 20 epochs with a learning rate of 1e-5. The maximum completion length was set to 2028 tokens for both the SFT and RL stages.

\subsection{Prompt template}
Here, we present the prompt designs employed in our method. Specifically, to enable \method to adopt a "think-with-images" generation paradigm, we employ the prompt shown in Fig.~\ref{fig:prompt_think}. During the final stage of the data generation pipeline, when a high-level model is used to complete the CoT reasoning, we apply the prompt depicted in Fig.~\ref{fig:prompt_cot}. For the reward computation of answer in RL stage, as well as for benchmark evaluation, we utilize the prompt provided in Fig.~\ref{fig:prompt_judge}.
Finally, for inference on the REC and OVD tasks, we adopt the prompt templates shown in Fig.~\ref{fig:prompt_rec} and Fig.~\ref{fig:prompt_ovd}, respectively.
\input{Figs/prompt_think}
\input{Figs/prompt_cot}
\input{Figs/prompt_judge}
\input{Figs/prompt_rec}
\input{Figs/prompt_ovd}

\section{More visualization}\label{sec:visualization}
In Fig.~\ref{fig:supvis1},~\ref{fig:supvis2},~\ref{fig:supvis3}, and~\ref{fig:supvis4}, we present additional visualizations of \method. For each question-image pair, we show the depth map generated by the depth interpreter, highlight the focused regions (red bounding boxes) of \method during the interleaved image-text reasoning. Ground-truth annotations are also provided for comparison.

\input{Figs/supvis1}
\input{Figs/supvis2}
\input{Figs/supvis3}
\input{Figs/supvis4}

%% file: Tabs/tab_dataset.tex
\begin{table*}[]
    \centering
    \footnotesize
    \begin{tabular}{lcc}
    \hline
    \textbf{Domain}                     & \textbf{Source Dataset} & \textbf{Size}  \\ \hline
    Counting                            & Tallyqa~\cite{acharya2019tallyqa}          & 370            \\ \hline
    \multirow{2}{*}{General VQA}        & Flickr30k~\cite{flickr30k}        & 9782           \\
                                        & Visual7W~\cite{visual7w}              & 8266           \\ \hline
    \multirow{3}{*}{Relation Reasoning} & GQA~\cite{gqa}              & 9481           \\
                                        & Open images~\cite{openimages}      & 8959           \\
                                        & VSR~\cite{vsr}              & 3041           \\ \hline
    Fine-Grained Understanding          & Birds-200-2011~\cite{cub}   & 9782           \\ \hline
                                        & \textbf{Sum}     & \textbf{49681} \\ \hline
    \end{tabular}
    \caption{The overview of the SIF-50K dataset. The dataset spans four distinct domains and includes various source datasets. }\label{table: dataset}
\end{table*}

%% file: Figs/prompt_think.tex
\begin{figure}[]
    \centering
    \includegraphics[width=0.46\textwidth]{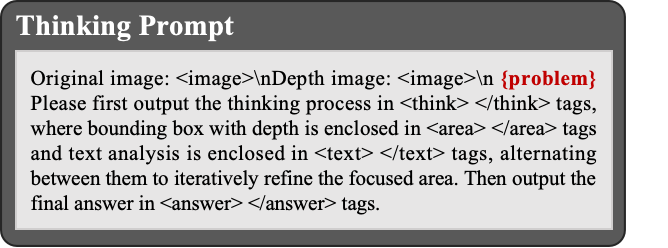}
    \caption{Prompt specifically crafted to guide the model in generating interleaved image-text reasoning chains, which is consistently appended during inference.}\label{fig:prompt_think}
\end{figure}

%% file: Figs/prompt_cot.tex
\begin{figure}[H]
    \centering
    \includegraphics[width=0.46\textwidth]{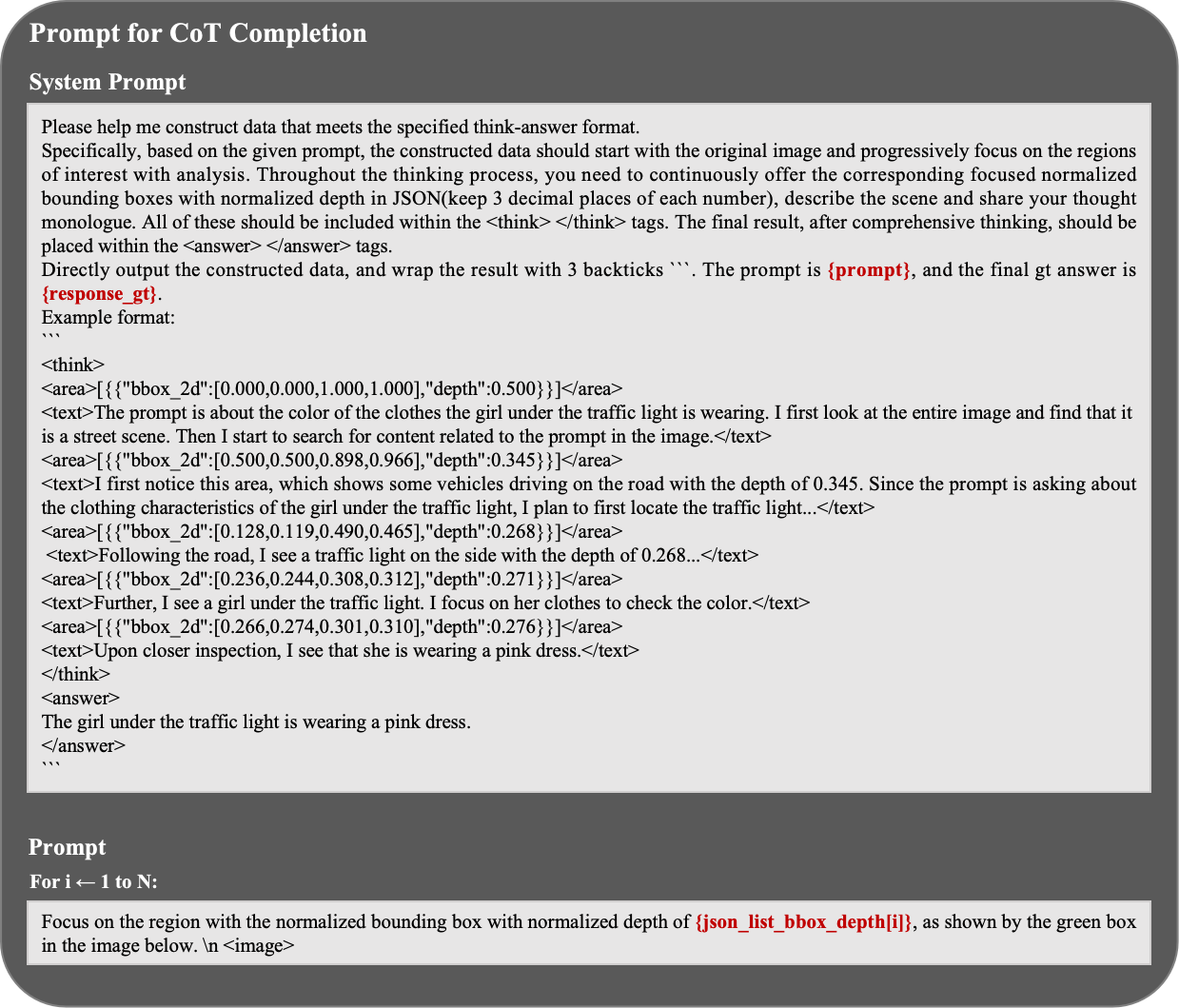}
    \caption{Prompt used for CoT generation, serving as the basis for constructing the SIF-50K dataset in our data generation pipeline.}\label{fig:prompt_cot}
\end{figure}

%% file: Figs/prompt_judge.tex
\begin{figure}[]
    \centering
    \includegraphics[width=0.46\textwidth]{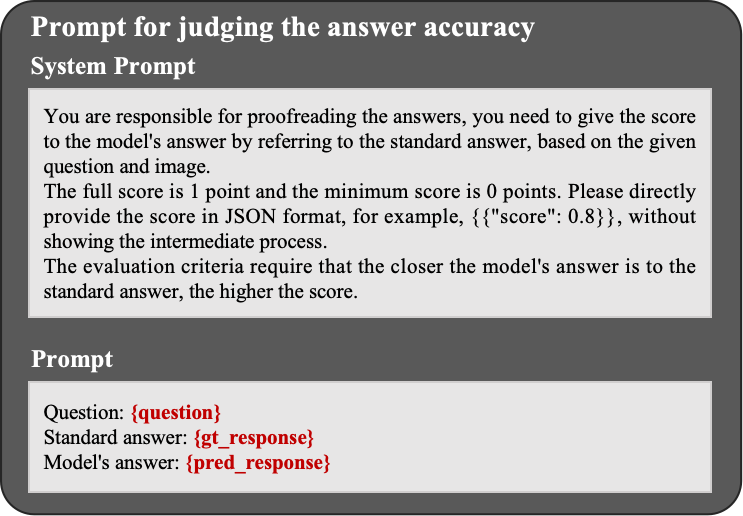}
    \caption{Prompt designed to assess the accuracy of model-generated answers against ground truth, used for computing $r_\text{ans}$ and for benchmark evaluation.}\label{fig:prompt_judge}
\end{figure}

%% file: Figs/prompt_rec.tex
\begin{figure}[H]
    \centering
    \includegraphics[width=0.46\textwidth]{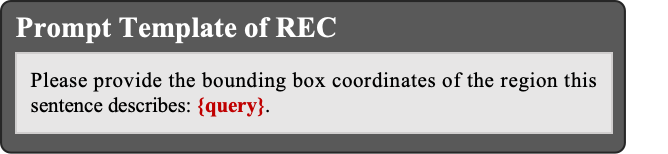}
    \caption{Prompt designed for evaluating model performance on the Referring Expression Comprehension (REC) task.}\label{fig:prompt_rec}
    \vspace{-0.8em}
\end{figure}

%% file: Figs/prompt_ovd.tex
\begin{figure}[H]
    \centering
    \includegraphics[width=0.46\textwidth]{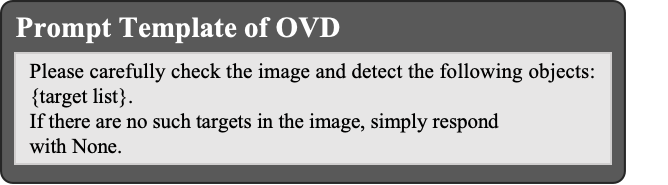}
    \caption{Prompt designed for evaluating model performance on the Open-Vocabulary Detection (OVD) task.}\label{fig:prompt_ovd}
    \vspace{-0.8em}
\end{figure}

%% file: Figs/supvis1.tex
\begin{figure*}[]
    \centering
    \includegraphics[width=1\textwidth]{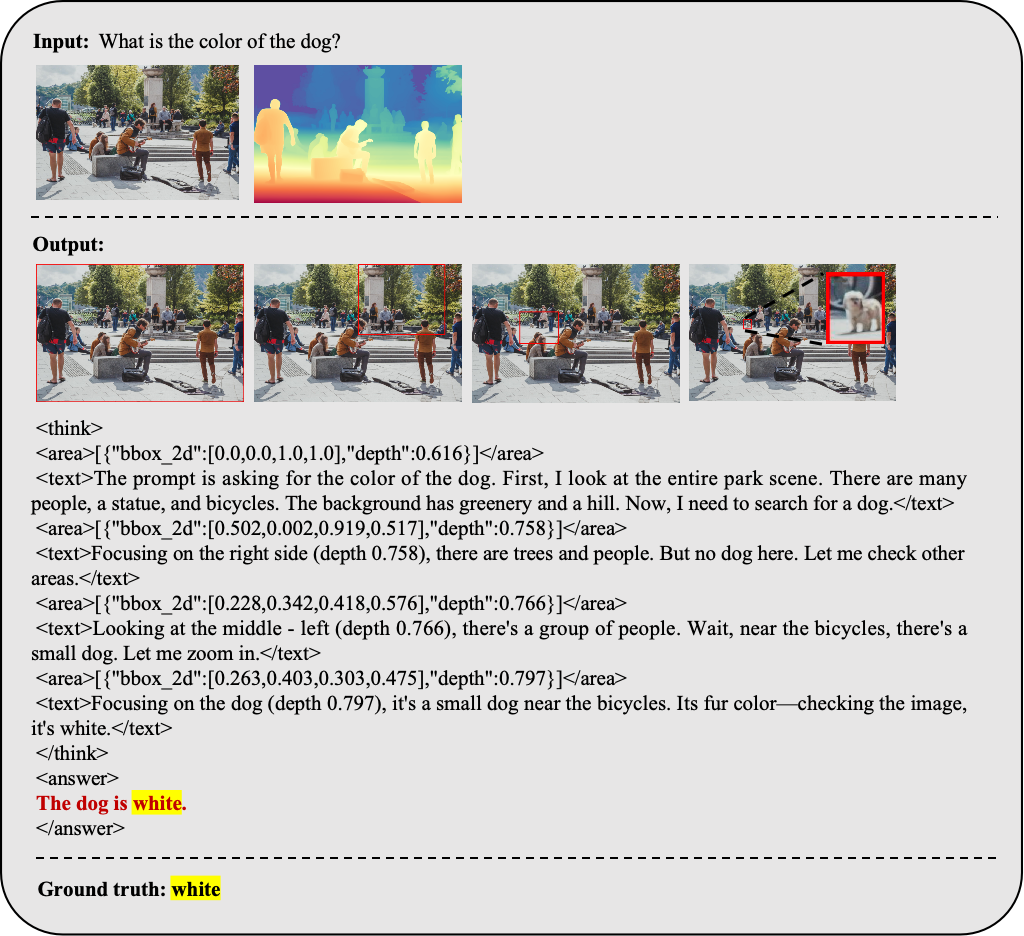}
    \caption{More visualization results of \method based on Qwen2.5-VL. We visualize the question, reasoning, ground truth, and highlight the areas of interest that the model focuses on during the reasoning process (red bounding boxes in the images).}\label{fig:supvis1}
\end{figure*}

%% file: Figs/supvis2.tex
\begin{figure*}[]
    \centering
    \includegraphics[width=1\textwidth]{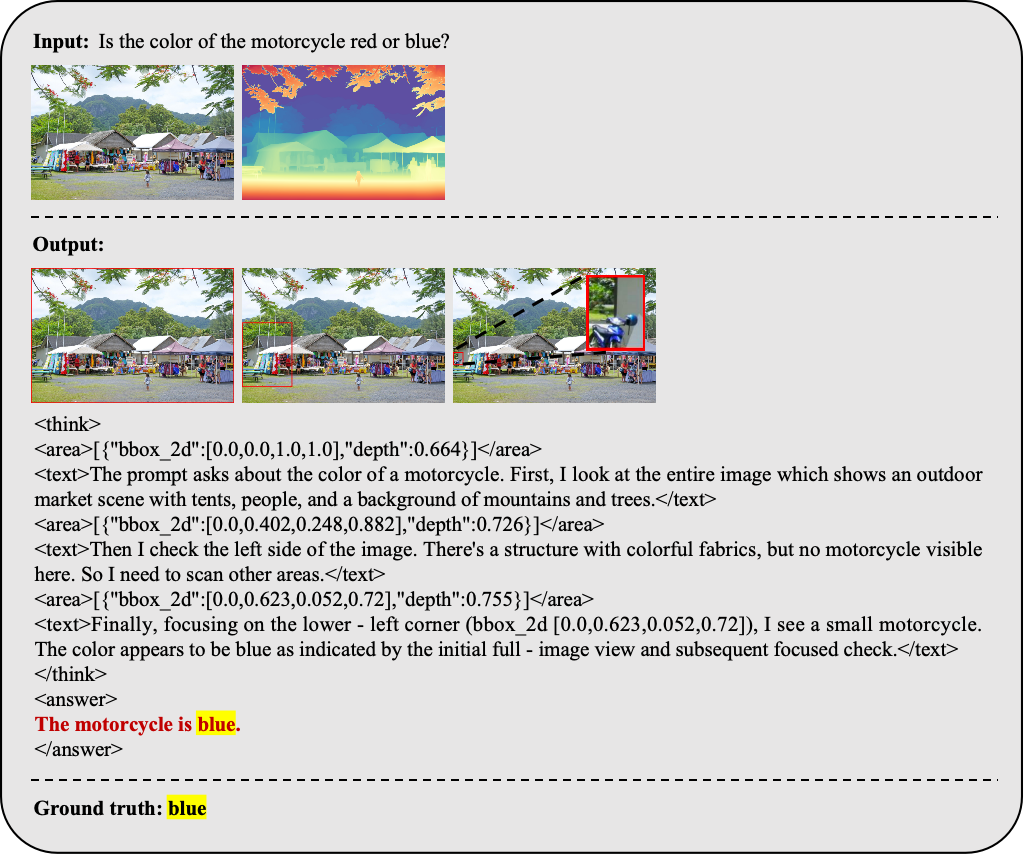}
    \caption{More visualization results of \method based on Qwen2.5-VL. We visualize the question, reasoning, ground truth, and highlight the areas of interest that the model focuses on during the reasoning process (red bounding boxes in the images).}\label{fig:supvis2}
\end{figure*}

%% file: Figs/supvis3.tex
\begin{figure*}[!htbp]
    \centering
    \includegraphics[width=1\textwidth]{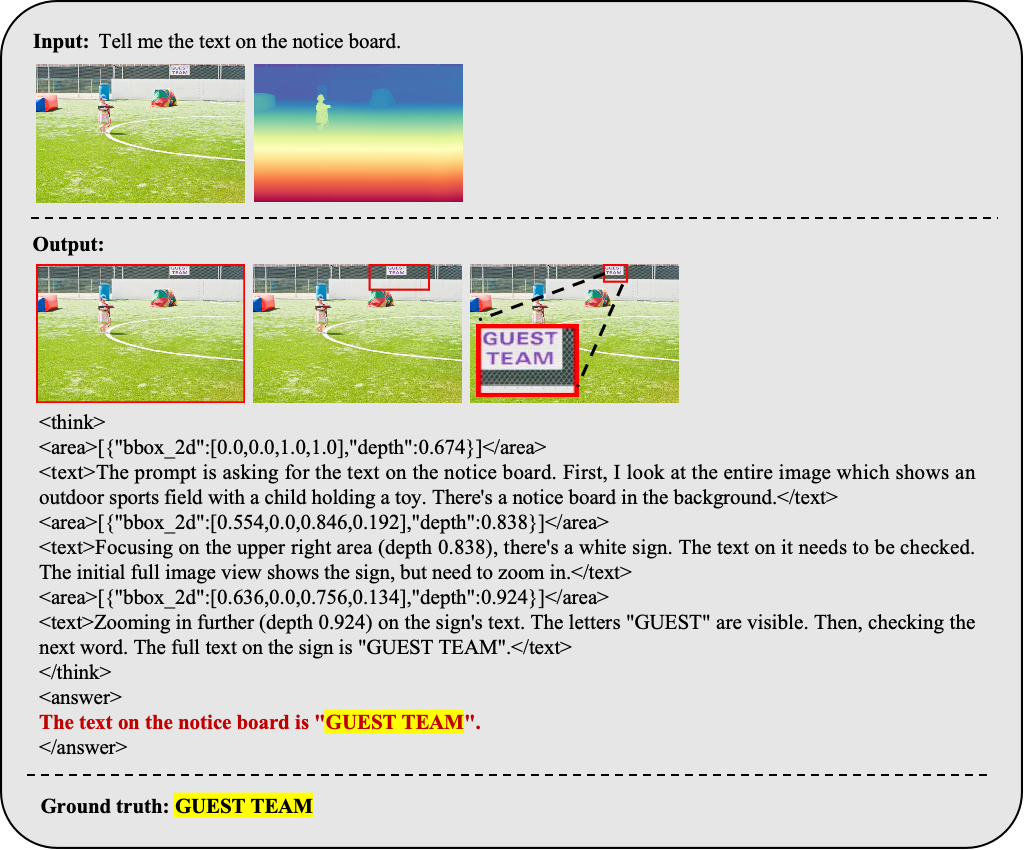}
    \caption{More visualization results of \method based on Qwen2.5-VL. We visualize the question, reasoning, ground truth, and highlight the areas of interest that the model focuses on during the reasoning process (red bounding boxes in the images).}\label{fig:supvis3}
\end{figure*}

%% file: Figs/supvis4.tex
\begin{figure*}[]
    \centering
    \includegraphics[width=1\textwidth]{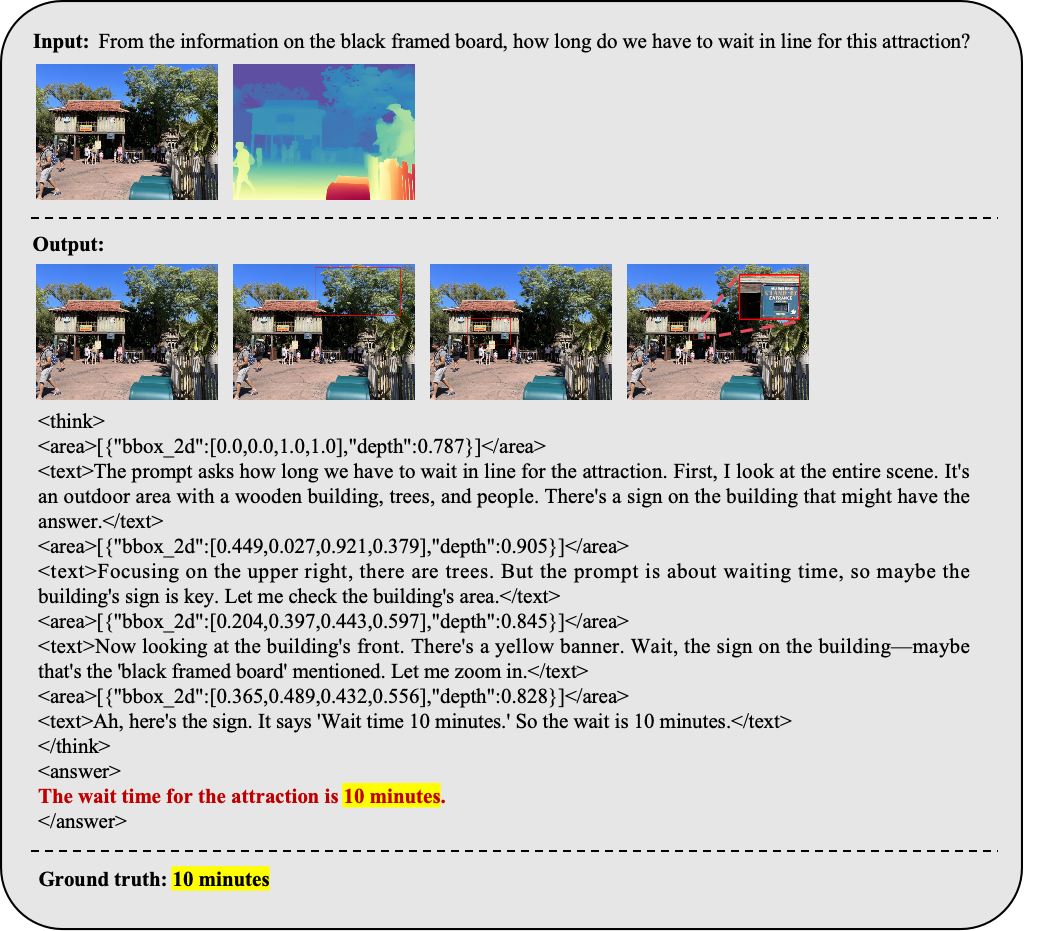}
    \caption{More visualization results of \method based on Qwen2.5-VL. We visualize the question, reasoning, ground truth, and highlight the areas of interest that the model focuses on during the reasoning process (red bounding boxes in the images).}\label{fig:supvis4}
\end{figure*}